\pdfoutput=1

\documentclass[11pt]{article}

\usepackage[review]{autoeval2021}

\usepackage{times}
\usepackage{latexsym}
\usepackage{enumitem}
\usepackage[symbol]{footmisc}
\usepackage[T1]{fontenc}

\usepackage[utf8]{inputenc}

\usepackage{microtype}

\usepackage{subcaption}
\usepackage{booktabs}
\usepackage{tabularx}
\usepackage{arydshln}
\usepackage{multirow}
\usepackage{graphicx}
\usepackage{array, makecell}
\usepackage{amssymb}

\title{User Response and Sentiment Prediction for Automatic Dialogue Evaluation}

\author{
Sarik Ghazarian
\thanks{~ Work done while Sarik Ghazarian was an intern at Amazon Alexa AI},\textsuperscript{\rm 1}
Behnam Hedayatnia,\textsuperscript{\rm 2} 
Alexandros Papangelis,\textsuperscript{\rm 2} \\
\textbf{Yang Liu},\textsuperscript{\rm 2}
\textbf{Dilek Hakkani-Tur},\textsuperscript{\rm 2}\\
\textsuperscript{\rm 1}University of Southern California / Information Sciences Institute \\
\textsuperscript{\rm 2}Amazon Alexa AI \\
sarik@isi.edu, \{behnam, papangea, yangliud, hakkanit\}@amazon.com
}

\begin{document}
\maketitle
\section{Introduction}
\label{sec:introduction}
\vspace*{-1ex}

Automatic evaluation is beneficial for open-domain dialog system development. However, standard word-overlap metrics (BLEU, ROUGE) do not correlate well with human judgements of open-domain dialog systems~\cite{deriu2020survey, liu2016not}.

Recent works have attempted to develop automatic evaluation methods that learn how to assess the generated responses given dialog contexts from different aspects 
such as relevancy~\cite{tao2018ruber, ghazarian2019better, lan2020pone}, engagement~\cite{ghazarian2020predictive}, fluency~\cite{zhang2021deep, pang2020towards}, contradiction~\cite{pang2020towards, nie2020like}, overall rating~\cite{jiang2021towards} amongst others.

Recently,~\citet{mehri2020unsupervised} have shown the effectiveness of the idea of looking into the next user utterance as a proxy to measure the quality of the chatbot's generated responses. 
We follow their idea with the main difference of not limiting the potential user follow-up utterances to predefined templates. 
We propose to use the sentiment of the next user utterance for turn or dialog level evaluation. Specifically we propose three methods: one that predicts the next sentiment directly, and two others that predict the next user utterance using an utterance or a feedback generator model and then classify its sentiment.

In contrast to most existing automatic evaluators that focus purely on text-based chatbots, 
we evaluate our proposed methods on both written and spoken dialog datasets.
Our contributions include: 
\begin{itemize}
\vspace{-0.03in}
    \item a next user response and feedback generator and next user sentiment estimator to evaluate open-domain dialogue systems at both dialog and turn levels,
\vspace{-0.03in}
    \item leveraging turn positions to better estimate dialog level scores,
\vspace{-0.03in}
    \item outperforming existing automatic evaluation metrics on both written and spoken open-domain dialogue datasets,
\vspace{-0.03in}
    \item an amended version of the DSTC9 dataset with turn level annotations denoted as ConTurE (https://github.com/alexa/conture).
\end{itemize}

\section{Dialog datasets}
\vspace*{-1ex}
\subsection{Spoken dialogs}
Given the limited work around evaluating spoken dialogs, we collect interactions between a spoken conversational agent and real users (denoted as~\textit{Real user interactions}). At the end of each interaction, the user is asked to leave a rating and a free-form feedback. In addition to interactions collected via real users, we also recruited paid users who were instructed to converse with a chatbot (denoted as~\textit{Paid user interactions}). 

In this work, we study the relationship between annotator (3P) and real user (1P) evaluation. We define 1P as users who have rated their own interactions with a conversational agent and 3P as annotators who rate other user's interactions.
We take a sample of ~\textit{Real user interactions} and annotate them at turn level. Given a complete interaction, an internal annotator will annotate each system response either as 1 or 0, where 1 indicates there is no issue with the response and vice versa for 0. Additionally, we ask the internal annotator to give a rating of 1-5 for the interaction. We denote the annotated data as~\textit{RUI} (687 interactions, average 11.7 turns). 
The same annotation was also performed on ~\textit{Paid user interactions} representing ~\textit{PUI} data (87 interactions, average 14.5 turns).

\subsection{Written dialogs}
We also annotate a set of written dialogs from the Interactive Evaluation of Dialog track of the Dialog State Tracking Challenge 9 (DSTC9)~\cite{gunasekara2020overview}. 
Authors asked AMT workers to rate 2200 interactions on multiple dialog-level parameters.
For turn level annotation, we sampled 119 dialogs with an average length of 8 turns resulting overall 1006 turns, asked  AMT workers to rate the overall impression they got from the chatbot's response with a score of either 0, 1, or 2 representing dislike, neutral, and like respectively. We denote this dataset as ConTurE (Conversational Turns Evaluation).

We also leverage FED dataset collected by~\citet{mehri2020unsupervised} including 125 dialog-level evaluations judged by five AMT workers.

\section{Feature analysis for dialog evaluation}
\label{sec:feature}
\vspace*{-1ex}
We obtained following findings by performing an analysis on turn and dialog level human annotation:
\vspace{-0.2in}

\begin{itemize}
    \itemsep -0.5ex
    \item Mean aggregation of turn level human annotation (w/o issues) is reasonably correlated with 3P dialog level ratings, but not 1P ones. 3P annotators had access to the whole dialog and could re-read the context while 1P can forget what happened in the dialog as it is spoken. 
    
    \item Positions of the turns matter when predicting dialog level scores, e.g., turns in the latter part of the dialogs have more effect. 

    \item User utterance sentiment's mean aggregation shows good correlation with dialog level scores. For the speech data, we used sentiment scores predicted using both acoustic and textural information \cite{kim2020speech}. 
\end{itemize}
\vspace*{-2ex}
\section{Models}
\vspace*{-1ex}
\subsection{Baseline models}
We leverage a suite of open source baseline models from~\cite{yeh2021comprehensive}
including RUBER, BERT-RUBER, PONE, PredictiveEngagement and FED metrics.
Due to space limitations, we only present results for FED as the most relevant work.

\vspace{-0.05in}
\subsection{Next user sentiment prediction} 
Based on the feature analysis results, we propose to train models to predict the next user sentiment for turn and dialog level evaluation. 
We evaluate different setups for this. 
\vspace{-0.08in}

\begin{itemize}
    \itemsep -0.5ex
    \item NUQ: Train a BERT-based classifier for turn quality prediction using dialogue context with different reference labels: 3P system turn annotation,  next user sentiment (that's automatically generated as described above), and whether or not the user stopped the interaction after that turn. 

    \item NUG: Generate the next user utterance and then obtain its sentiment scores. We finetuned a GPT2-medium model on the spoken data, and achieved a perplexity of 12.8. We used the out of the box DialoGPT for the text data.

    \item NUF: Generate a feedback style utterance for the next user turn, and then obtain the sentiment of the feedback. We used the actual feedback from the \textit{Real user interactions} that is more indicative of the dialog quality to train a DialoGPT model for feedback generation. The text-based sentiment classifier used in this and the above setting is from HuggingFace. 
\end{itemize}

\vspace{-0.17in}
Once we have turn level predictions, we aggregate them using unweighted or weighted average to obtain dialog level scores.  

\vspace{-0.05in}
\section{Results}
\vspace*{-1ex}
Table~\ref{tab:results} shows the automatic evaluation results on different datasets, two speech based
(RUI and PUI), and two text datasets (ConTurE and FED).
FED performs poorly on the spoken dialog dataset for both turn and dialog level annotation. The best results obtained are using our proposed model to predict next user utterance quality. 
Using sentiment of the generated next user utterance or feedback do not achieve as good correlation, possibly because of  poor generation or the fact that the generated utterance is not good indicator of the previous system response quality. 
We also observe some differences among different data sets. 
It is worth noting that the correlation between the 3P response quality annotation and the next user sentiment prediction is not high. We hypothesize this may be related to the noise of the 3P turn level annotation and will further study this in our future work.

\begin{table} 
    \centering
    \begin{small}
    \begin{tabular}{|c|c|c|c|c|c|} \hline
       \multicolumn{2}{|c|}{}  & FED & NUG & NUF & NUQ* \\ \hline
    \multirow{3}{*}{RUI} &   3p turn  & -0.006 & 0.05 & 0.07 & \textbf{0.17} \\ \cline{2-6}
    & 3p dialog & 0.22 & 0.04 & 0.20 & \textbf{0.27} \\ \cline{2-6}
    & 1p dialog & 0.10 & -0.05 & 0.13 & \textbf{0.36} \\ \hline
    \multirow{2}{*}{PUI} & 3p turn & -0.03& 0.03 & 0.03 &  \textbf{0.24}\\ \cline{2-6}
    & 3p dialog & 0.21& 0.13 & 0.04 & \textbf{0.24} \\ \hline
    \multirow{2}{*}{ConTurE} & 3p turn & 0.11 & 0.12 & 0.02& \textbf{0.19}\\ \cline{2-6} 
    & 3p dialog & 0.13 & 0.10 & 0.10& \textbf{0.27}\\ \hline
    FED data & 3p dialog & 0.22 & 0.24 & 0.19 & \textbf{0.38}\\ \hline 
    \end{tabular}
    \caption{Pearson correlation results using FED baseline and our methods on different datasets. For NUQ*, we show the best results from different configurations (the training labels used and position weighting when aggregating turn level predictions for dialog level ratings).}
    \label{tab:results}
    \end{small}
\vspace{-1em}
\end{table}

\vspace{-0.05in}
\section{Conclusion}
\vspace*{-1ex}
In this work we explored the idea of looking into future turns and their sentiments to evaluate the quality of the generated responses. According to our findings, predicting users' next utterance sentiment scores has a positive impact on the accuracy of the evaluation metrics.

\bibliography{anthology,autoeval2021}
\bibliographystyle{acl_natbib}

\end{document}